\documentclass{article}
\pdfoutput=1

\PassOptionsToPackage{numbers,compress}{natbib}
\usepackage[workshop,final]{neurips_2026}

\usepackage[utf8]{inputenc}
\usepackage[T1]{fontenc}
\usepackage{hyperref}
\usepackage{url}
\usepackage{graphicx}
\usepackage{booktabs}
\usepackage{amsmath}
\usepackage{amssymb}
\usepackage{amsfonts}
\usepackage{amsthm}
\usepackage{mathtools}
\usepackage{microtype}
\usepackage{xcolor}
\usepackage{xspace}
\usepackage{enumitem}
\usepackage{multirow}
\usepackage[title]{appendix}

\newtheorem{proposition}{Proposition}

\newcommand{\tr}{\operatorname{tr}}
\newcommand{\method}{\textsc{Moment-DP}\xspace}
\newcommand{\Ddyn}{D_{\mathrm{dyn}}}

\title{
MOMENT-DP:
Momentum-Aware Evaluation of Correlated Noise
for Differentially Private Training
}

\author{
Sunny Gupta \\
Google Cloud \\
\texttt{sunnyguptaa@google.com}
\And
Raja Viswanadha \\
Google Cloud \\
\texttt{rajaviswanadha@google.com}
}

\begin{document}

\maketitle


\begin{abstract}

Temporally correlated noise can substantially improve the utility of
differentially private optimization, yet candidate mechanisms are commonly
designed or evaluated using only partial information about the optimization
process, such as workload structure, learning-rate schedules, or local
curvature. We study whether persistent optimizer state provides additional
information that can change the relative utility of privacy-valid
correlated-noise mechanisms.

We introduce \method, an optimizer-state-aware framework that models how
temporally correlated privacy noise propagates through momentum-based
optimization. For locally quadratic objectives, \method expresses
privacy-induced final loss through an optimizer transfer operator that jointly
captures curvature, learning-rate scheduling, momentum, and temporal noise
covariance. NoiseCurve's curvature-aware quadratic criterion is recovered
exactly as the zero-momentum, constant-learning-rate special case.

Changing only momentum while holding curvature, schedule, privacy calibration,
and candidate mechanisms fixed induces genuine mechanism-ranking reversals.
The effect persists outside exact quadratic dynamics in per-example-clipped
private logistic regression, while remaining unresolved in the neural regime
we study. These results establish optimizer state as an under-modeled,
decision-relevant axis in evaluating temporally correlated privacy noise.

\end{abstract}


\section{Introduction}

Differentially private stochastic gradient descent (DP-SGD) protects training
examples by clipping individual gradients and adding calibrated Gaussian noise
\citep{abadi2016deep}. Rather than injecting independent perturbations at every
iteration, DP-FTRL and matrix-factorization (MF) mechanisms correlate privacy
noise over time and can substantially improve privacy--utility trade-offs
\citep{kairouz2021practical,denisov2022improved,choquette2023banded}.

This raises a basic question: \emph{how should temporal privacy-noise
covariances be ranked for learning?} Classical MF criteria operate largely at
the workload level. More recent approaches incorporate additional optimization
information: learning-rate-aware MF explicitly models the schedule
\citep{kalinin2026learning}, while NoiseCurve incorporates local curvature
\citep{gu2025noisecurve}. We investigate another information axis that remains
under-modeled in this mechanism-evaluation lineage:
\textbf{persistent optimizer state}.

Momentum is not merely a learning-rate rescaling. Earlier gradients and
privacy perturbations repeatedly influence later updates through the momentum
buffer. Consequently, two privacy-valid covariances can exchange ordering as
momentum changes even when curvature, learning-rate schedule, privacy
calibration, and the candidate mechanisms remain fixed.

We formalize this interaction using a state-space representation of private
momentum SGD. For locally quadratic optimization, temporally correlated
privacy noise with covariance $K$ contributes expected final loss

\begin{equation}
\Ddyn
=
\frac{\sigma^2}{2}
\tr\!\left[
H G(K\otimes I_d)G^\top
\right],
\label{eq:ddyn_intro}
\end{equation}

where the transfer operator $G$ depends on curvature, the full learning-rate
schedule, momentum, and the optimization horizon.

A central result places Eq.~\eqref{eq:ddyn_intro} directly in the context of prior
work: NoiseCurve's exact quadratic criterion is precisely its
zero-momentum, constant-learning-rate specialization. Thus, our question is
not whether curvature matters---NoiseCurve already establishes that it
does---but whether persistent optimizer state provides additional information
needed to rank privacy-valid temporal covariances.

\textbf{Contributions.}
We make three contributions:
(i) an exact optimizer-state-aware transfer formulation for temporally
correlated privacy noise under quadratic momentum dynamics;
(ii) an exact reduction to NoiseCurve's quadratic objective at zero momentum
and constant learning rate, numerically verified across 576 configurations;
and
(iii) controlled and nonlinear evidence that changing only optimizer state can
reverse the relative utility of privacy-valid mechanisms, together with an
explicit negative neural result.


\begin{figure}[t]
\centering
\includegraphics[width=\textwidth]{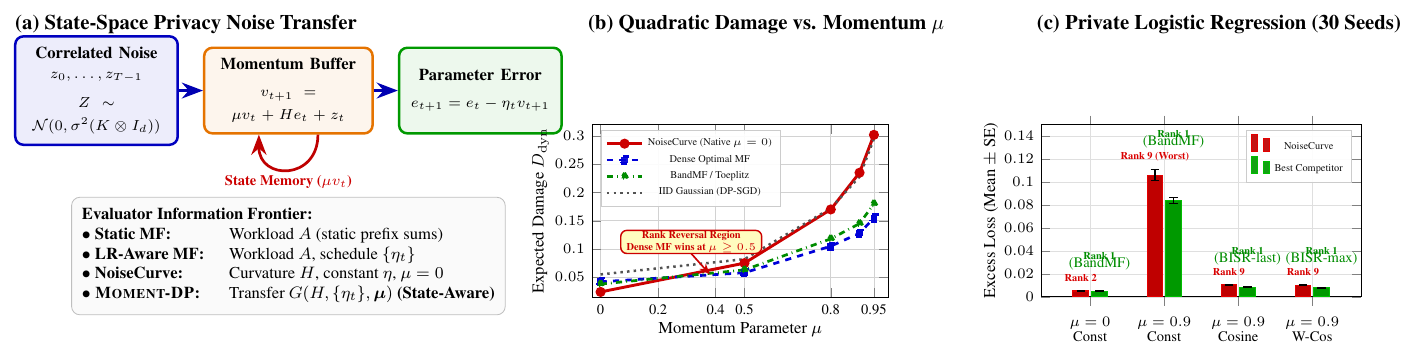}
\vspace{-3mm}

\caption{
\textbf{\method overview.}
\textbf{(a)} Temporally correlated privacy perturbations are recursively
filtered through persistent momentum state before affecting parameter error.
\textbf{(b)} Holding curvature, learning rate, privacy calibration, and
candidate covariances fixed, changing only momentum changes the oracle-best
mechanism: NoiseCurve is optimal at $\mu=0$, while Dense MF is optimal at
every tested $\mu\ge0.5$.
\textbf{(c)} In per-example-clipped private logistic regression
(30 paired seeds), NoiseCurve changes from rank $2/9$ at zero momentum to
rank $9/9$ in each of the three tested momentum conditions.
}
\label{fig:overview}

\vspace{-2mm}
\end{figure}


\section{MOMENT-DP}

\subsection{State-space formulation}

Consider
$f(\theta)=\frac12(\theta-\theta^\star)^\top
H(\theta-\theta^\star)$, $H\succeq0$, and let
$e_t=\theta_t-\theta^\star$. The stacked privacy perturbations satisfy
\[
Z=[z_0^\top,\ldots,z_{T-1}^\top]^\top
\sim\mathcal N(0,\sigma^2(K\otimes I_d)),
\]
where $K\in\mathbb R^{T\times T}$ is the temporal covariance. Momentum SGD
obeys
\begin{equation}
v_{t+1}=He_t+\mu v_t+z_t,
\qquad
e_{t+1}=e_t-\eta_t v_{t+1}.
\label{eq:momentum}
\end{equation}

With $s_t=[e_t^\top,v_t^\top]^\top$, the dynamics are
$s_{t+1}=A_ts_t+B_tz_t$, where
\[
A_t=
\begin{bmatrix}
I-\eta_tH & -\eta_t\mu I\\
H & \mu I
\end{bmatrix},
\qquad
B_t=
\begin{bmatrix}
-\eta_tI\\
I
\end{bmatrix}.
\]
Let $G$ map the stacked perturbations to final parameter error,
$e_T^{(z)}=GZ$. Then
\[
\operatorname{Cov}(e_T^{(z)})
=
\sigma^2G(K\otimes I_d)G^\top,
\]
and therefore
\begin{equation}
D_{\mathrm{dyn}}
=
\frac{\sigma^2}{2}
\operatorname{tr}
\!\left[
HG(K\otimes I_d)G^\top
\right].
\label{eq:ddyn}
\end{equation}

\begin{proposition}[Exact quadratic damage]
\label{prop:exact}
Under fixed quadratic dynamics, predetermined learning rates, momentum
$\mu$, and temporal covariance $K$, Eq.~\eqref{eq:ddyn} equals the exact
expected final objective contribution caused solely by privacy noise.
\end{proposition}

The proof is given in Appendix~\ref{app:proofs}. Since \method is exact in
this domain, quadratic ranking accuracy is calibration rather than independent
predictive evidence.

\subsection{NoiseCurve as an exact special case}
\label{sec:noisecurve}

For $\mu=0$ and constant $\eta$, Hessian eigenmode $i$ satisfies
\begin{equation}
e_{t+1}^{(i)}
=
(1-\eta\lambda_i)e_t^{(i)}-\eta z_t^{(i)}.
\label{eq:eigenmode}
\end{equation}
Its temporal transfer vector therefore obeys
$g_i=-\eta V_{i,:}$, where
$V_{ij}=(1-\eta\lambda_i)^{T-j-1}$.
Writing $M=\operatorname{diag}(\lambda_1,\ldots,\lambda_d)$,
$X=C^\top C$, and $K=X^{-1}$ gives
\begin{equation}
D_{\mathrm{dyn}}
=
\frac{\eta^2\sigma^2}{2}
\operatorname{tr}\!\left(X^{-1}V^\top MV\right),
\label{eq:noisecurve}
\end{equation}
which is exactly NoiseCurve's quadratic final-loss criterion
\citep{gu2025noisecurve}. Importantly, the identity is mode-wise:
the full $G\in\mathbb R^{d\times dT}$ is not identified with
$V\in\mathbb R^{d\times T}$.

We verify this identity over 576 configurations; the worst relative numerical
discrepancy is $9.7\times10^{-16}$. Full derivation and validation are in
Appendices~\ref{app:proofs} and~\ref{app:validation}.


\paragraph{Momentum-dependent rankings.}
For mechanisms $a$ and $b$,
\[
\Delta(\mu)
=
\frac{\sigma^2}{2}
\operatorname{tr}\!\left[
HG_\mu((K_a-K_b)\otimes I_d)G_\mu^\top
\right].
\]
If $\Delta(\mu_1)\Delta(\mu_2)<0$, their oracle ordering reverses.
Any criterion independent of $\mu$ assigns the same ordering at both
momentum values and therefore cannot be correct at both.


\section{Experiments and Results}

We evaluate IID Gaussian noise, dense MF, BandMF, Toeplitz MF, BISR,
learning-rate-aware MF, and NoiseCurve under matched single-participation
privacy semantics. Controlled experiments use
$(\varepsilon,\delta)=(4,10^{-5})$.
Learning-rate-aware MF and NoiseCurve are independently reproduced and
validated. To reduce outcome-informed selection, analytic crossover discovery
is separated from downstream evaluation:
cases are selected analytically, serialized and SHA-256 frozen, and then
evaluated using an independent NumPy simulator.
Full mechanism, privacy, solver, and protocol details appear in
Appendices~\ref{app:validation}--\ref{app:blind}.


\subsection{Controlled momentum crossovers}

All 12 frozen competitive quadratic crossover cases are independently
confirmed. Because \method is exact in this domain, the result establishes
the existence and reproducibility of the crossovers rather than independent
predictive superiority.

Holding curvature and learning-rate schedule fixed while varying
$\mu\in\{0,0.5,0.9,0.95\}$, the oracle-best mechanism changes in $7/9$
tested curvature--schedule slices.
The faithfully reproduced learning-rate-aware ranking changes in $0/7$ of
those seven momentum-driven crossover slices.

Across the broader exact grid, pairwise ranking accuracy is $58.0\%$ for
static MeanSE, $60.8\%$ for learning-rate-aware MF, and $75.8\%$ for a
curvature-and-schedule score evaluated with momentum forced to zero.
Full results are in Appendix~\ref{app:quadratic}.


\subsection{Direct NoiseCurve intervention}

NoiseCurve is optimized once in its native condition
($\mu=0$, constant learning rate).
We then hold $H$, $\eta$, privacy calibration, and the candidate pool fixed,
and change only momentum.

\begin{table}[t]
\centering
\small

\caption{
Changing only momentum changes the oracle-best mechanism.
}
\label{tab:intervention}

\begin{tabular}{rlll}
\toprule
$\mu$
& Oracle best
& NC score
& \method \\
\midrule
0.00 & NoiseCurve & correct & correct \\
0.50 & Dense MF & wrong & correct \\
0.80 & Dense MF & wrong & correct \\
0.90 & Dense MF & wrong & correct \\
0.95 & Dense MF & wrong & correct \\
\bottomrule
\end{tabular}

\end{table}

NoiseCurve correctly wins in its native zero-momentum regime.
Its exact criterion contains no momentum parameter, so its score does not
change under this intervention.
The result should therefore not be interpreted as a defect of NoiseCurve;
instead, it demonstrates that changing an optimizer-state variable absent from
the zero-momentum criterion can change the preferred mechanism.


\subsection{Nonlinear private logistic regression}
\label{sec:logistic}

We next test whether the distinction survives outside exact quadratic
dynamics.
The private logistic-regression experiment uses $d=16$, $N=4096$,
$T=128$ disjoint single-participation shards, real per-example clipping,
30 paired seeds, and $(\varepsilon,\delta)=(4,10^{-5})$.
Unlike the quadratic setting, clipping, stochastic gradients, curvature drift,
and mechanism-dependent trajectories are present.

\begin{table}[t]
\centering
\small

\caption{
NoiseCurve under optimizer interventions in private logistic regression
(mean $\pm$ SE over 30 paired seeds).
}
\label{tab:logistic}

\begin{tabular}{lrrl}
\toprule
Condition
& NC rank
& NC excess loss
& Best excess loss \\
\midrule

$\mu=0$, constant
& 2/9
& $0.00505\pm0.00015$
& $0.00497\pm0.00017$ \\

$\mu=.9$, constant
& 9/9
& $0.10590\pm0.00487$
& $0.08390\pm0.00242$ \\

$\mu=.9$, cosine
& 9/9
& $0.01048\pm0.00039$
& $0.00840\pm0.00023$ \\

$\mu=.9$, warmup-cos.
& 9/9
& $0.00999\pm0.00043$
& $0.00785\pm0.00021$ \\

\bottomrule
\end{tabular}

\end{table}

NoiseCurve is near-optimal in its native zero-momentum regime, but becomes the
worst of nine candidate mechanisms in all three tested momentum conditions.
The intervention was specified before observing the empirical outcomes.

Across the broader logistic study, the local \method approximation improves
statistically resolved pairwise ranking accuracy from $91.8\%$ to $95.1\%$
and reduces mean relative selection regret from $4.41\%$ to $3.53\%$,
while top-1 agreement remains unchanged at $44.4\%$.
We therefore claim improved mechanism-ordering information rather than
reliable universal top-1 selection.


\section{Neural Boundary}

A five-seed CIFAR-10 CNN pilot does not establish reliable neural mechanism
ranking transfer.
Across four optimizer conditions, best--worst mechanism spreads are only
$0.29$--$0.48$ percentage points and the local \method approximation does
not consistently identify the empirical winner.

We therefore report \textbf{no confirmed neural ranking crossover}.
The experiment uses a short $T=100$ single-participation trajectory and
should be interpreted as a boundary test of the present local approximation,
not as evidence that optimizer state is irrelevant to neural private training.
Architecture, calibration, absolute results, and possible contributing factors
are reported in Appendix~\ref{app:cnn}.


\section{Conclusion}

\method shows that the utility of temporally correlated privacy noise can
depend on the optimizer state that consumes it.
Its quadratic transfer formulation exactly recovers NoiseCurve at zero
momentum and constant learning rate, while exposing mechanism-ranking
reversals when momentum changes.
Controlled experiments and per-example-clipped private logistic regression
confirm that this distinction can be decision-relevant, whereas neural
transfer remains unresolved.
These results motivate evaluating privacy-mechanism selection and optimizer
configuration jointly.

\newpage


\label{mainend}


\bibliographystyle{plainnat}
\bibliography{refs}


\newpage
\begin{appendices}
\section{Proofs and Additional Theory}
\label{app:proofs}

\subsection{Expanded proof of Proposition~\ref{prop:exact}}

For $t<T$, define

\[
\Phi_{T,t}
=
A_{T-1}\cdots A_t,
\]

with empty-product convention

\[
\Phi_{T,T}=I_{2d}.
\]

Let

\[
P=[I\;0]
\]

extract the parameter-error block.
Unrolling the state recursion gives the privacy-noise component

\[
s_T^{(z)}
=
\sum_{t=0}^{T-1}
\Phi_{T,t+1}B_tz_t.
\]

Applying $P$,

\[
e_T^{(z)}
=
\sum_{t=0}^{T-1}
P\Phi_{T,t+1}B_tz_t.
\]

Define

\[
G_t
=
P\Phi_{T,t+1}B_t,
\qquad
G=
[G_0,\ldots,G_{T-1}].
\]

Then

\[
e_T^{(z)}=GZ.
\]

Since

\[
Z
\sim
\mathcal N
\left(
0,\sigma^2(K\otimes I_d)
\right),
\]

we have

\[
\mathbb E
\left[
e_T^{(z)}
e_T^{(z)\top}
\right]
=
\sigma^2
G(K\otimes I_d)G^\top.
\]

Therefore,

\[
\begin{aligned}
\mathbb E
\left[
\frac12
e_T^{(z)\top}
He_T^{(z)}
\right]
&=
\frac12
\tr
\left(
H
\mathbb E[
e_T^{(z)}
e_T^{(z)\top}
]
\right)
\\
&=
\frac{\sigma^2}{2}
\tr
\left[
HG(K\otimes I_d)G^\top
\right].
\end{aligned}
\]

This is Eq.~\eqref{eq:ddyn}.

\subsection{Expanded derivation of Eq.~\eqref{eq:noisecurve}}

For $\mu=0$ and constant $\eta$,
Eq.~\eqref{eq:eigenmode} unrolls mode by mode as

\[
e_T^{(i,z)}
=
-\eta
\sum_{j=0}^{T-1}
(1-\eta\lambda_i)^{T-j-1}
z_j^{(i)}.
\]

Define

\[
V_{ij}
=
(1-\eta\lambda_i)^{T-j-1}
\]

and the temporal transfer vector

\[
g_i
=
[g_{i0},\ldots,g_{i,T-1}].
\]

Then

\[
g_i=-\eta V_{i,:}.
\]

The full operator
$G\in\mathbb R^{d\times dT}$
is obtained by combining these mode-wise temporal transfer vectors in the
Hessian eigenbasis; the full block operator $G$ is therefore not identified
directly with the $d\times T$ matrix $V$.

Let

\[
M
=
\operatorname{diag}
(\lambda_1,\ldots,\lambda_d),
\qquad
X=C^\top C.
\]

Since

\[
K=C^{-1}C^{-\top}=X^{-1},
\]

\[
\begin{aligned}
\Ddyn
&=
\frac{\sigma^2}{2}
\sum_{i=1}^{d}
\lambda_i g_iX^{-1}g_i^\top
\\
&=
\frac{\eta^2\sigma^2}{2}
\sum_{i=1}^{d}
\lambda_i
V_{i,:}
X^{-1}
V_{i,:}^\top
\\
&=
\frac{\eta^2\sigma^2}{2}
\tr
\left(
X^{-1}V^\top MV
\right).
\end{aligned}
\]

This is exactly NoiseCurve's quadratic objective.

\subsection{Trajectory-weighted extension}

If utility over the optimization trajectory is of interest rather than only
final loss, let $G^{(t)}$ map the full perturbation sequence to $e_t$.

A corresponding score is

\[
D_{\mathrm{traj}}
=
\sum_{t=1}^{T}
w_t
\frac{\sigma^2}{2}
\tr
\left(
H_t
G^{(t)}
(K\otimes I_d)
G^{(t)\top}
\right).
\]

We explored preregistered uniform weights, but final-state damage was more
discriminative in the controlled ranking tasks and is therefore used for
headline results.


\section{Prior-Work Reproduction and Numerical Validation}
\label{app:validation}

\subsection{Learning-rate-aware matrix factorization}

We implement the schedule-aware factorization of
\citet{kalinin2026learning}
directly from its published equations.
For schedule multipliers $\chi$, the prefix workload becomes

\[
A_\chi
=
A_1\operatorname{diag}(\chi).
\]

The implementation is validated in its constant-learning-rate limit against
the trusted Toeplitz strategy in JAX-Privacy, with machine-precision
agreement.

\subsection{NoiseCurve Problem P1}

NoiseCurve optimizes

\[
X^\star
=
\arg\min_X
\tr(QX^{-1}),
\qquad
Q=V^\top MV,
\]

subject to

\[
\operatorname{diag}(X)=\mathbf 1,
\qquad
X\succ0,
\]

and a prescribed band structure.

Our primary implementation parameterizes the free symmetric band entries of
$X$ and fixes its diagonal exactly.
Optimization uses Adam with learning rate $0.02$.
The free entries are initialized as

\[
0.05\,\mathcal N(0,1)
\]

using random seed $0$.

Positive definiteness is enforced using a floor $10^{-3}$.
When

\[
\lambda_{\min}(X)<10^{-3},
\]

the objective receives the additional quadratic penalty

\[
10^6
\left(
10^{-3}-\lambda_{\min}(X)
\right)^2.
\]

Default validation runs use 2000 optimizer steps.
Dedicated solver-validation runs use 3000 steps, while the logistic-regression
NoiseCurve mechanism uses 1500 steps.

We independently cross-check the implementation using SciPy L-BFGS-B with

\[
\texttt{maxiter}=2000,
\qquad
\texttt{ftol}=10^{-14},
\qquad
\texttt{gtol}=10^{-10},
\]

for

\[
T\in\{8,16,32\},
\qquad
b\in\{2,4\}.
\]

Four of six tested configurations agree within $0.01\%$ objective value.
The two $T=32$ configurations differ by approximately $0.9\%$ and $4.7\%$,
respectively; both returned matrices remain positive definite.

These $T=32$ instances are used only for solver cross-validation and do not
enter the headline mechanism rankings.

Recovered strategies satisfy

\[
\frac{
\|C^\top C-X\|_F
}{
\|X\|_F
}
\approx
10^{-16}.
\]

During development, an error was detected in the independent validation
script: indefinite matrices were diagonally regularized without properly
accounting for violation of the positive-definite constraint.
The primary optimizer already contained the intended positive-definiteness
penalty.
The independent validation script was corrected before accepting scientific
results based on the reproduction.

\subsection{NoiseCurve--\method identity grid}

The 576-case identity grid uses

\[
T\in\{8,32,128\},
\]

six curvature spectra, and

\[
\eta\lambda_{\max}
\in
\{0.05,0.1,0.3,0.6\}.
\]

We compute:

\begin{enumerate}[leftmargin=1.5em]
    \item NoiseCurve's published quadratic objective;
    \item $\Ddyn$ using the implemented recurrence; and
    \item direct covariance propagation.
\end{enumerate}

All 576 configurations agree to relative error below $10^{-10}$.
The worst observed discrepancy is

\[
9.7\times10^{-16}.
\]


\section{Privacy Validation and Mechanism Catalog}
\label{app:privacy}

All headline controlled and logistic comparisons use the same declared privacy
unit, single-participation semantics, and
$(\varepsilon,\delta)$ target.

Candidate strategy matrices are independently checked and, where necessary,
normalized to sensitivity one before Gaussian calibration.
For the controlled single-participation setting, we use the analytic Gaussian
mechanism \citep{balle2018improving}.

Because sensitivity and privacy target are common across the mechanisms in
these comparisons, the scalar $\sigma^2$ factor is shared and therefore does
not alter exact quadratic rankings.

The competitive pool contains:

\begin{itemize}[leftmargin=1.5em]
    \item IID Gaussian;
    \item dense optimal MF;
    \item BandMF mean/max;
    \item Toeplitz mean/max;
    \item BISR variants;
    \item learning-rate-aware MF; and
    \item NoiseCurve.
\end{itemize}

BISR denotes the Banded Inverse Square Root construction
\citep{kalinin2026square}.
The strategies used in the controlled and logistic studies are exported from
the JAX-Privacy implementation with

\[
\texttt{min\_sep}=128,
\qquad
\texttt{max\_participations}=1,
\qquad
\texttt{num\_bands}=16.
\]

For the $T=128$ experiments, banded constructions use semi-bandwidth $b=16$.

An exploratory last-only stress mechanism produced extreme MeanSE/MaxSE and
was excluded from the competitive pool before the final comparative stage.


\section{Blind Crossover Protocol}
\label{app:blind}

The blind protocol separates analytic discovery from independent validation
to reduce outcome-informed case selection.

\paragraph{Discovery grid.}

The analytic search spans

\[
\eta\lambda_{\max}
\in
\{
0.0025,
0.005,
0.01,
0.02,
0.05,
0.10,
0.20,
0.40,
0.60,
0.80
\},
\]

momentum

\[
\mu
\in
\{
0,
0.5,
0.8,
0.9,
0.95,
0.99
\},
\]

and eight learning-rate schedule families:

\begin{itemize}[leftmargin=1.5em]
    \item constant;
    \item warmup-constant;
    \item cosine;
    \item warmup-cosine;
    \item step decay;
    \item exponential decay;
    \item front-loaded; and
    \item back-loaded.
\end{itemize}

The controlled objectives use six $d=16$ curvature-spectrum families,
including isotropic, condition numbers $10$, $100$, and $1000$, and two
additional stress spectra.

\paragraph{Competitive-case selection.}

Pairwise mechanism disagreements qualify for the primary registry when their
relative oracle margin exceeds $10\%$.

A secondary control registry uses a $5\%$ threshold.

Cases are balanced across optimizer and curvature conditions before any
downstream Monte Carlo outcome is examined.

\paragraph{Freezing and hashing.}

Selected cases are serialized into a machine-readable registry and hashed
using SHA-256 before independent simulation.

Twelve cases form the primary frozen crossover registry.

\paragraph{Independent validation.}

Validation uses a NumPy simulator implemented independently of the JAX/autodiff
code used to compute $\Ddyn$.

The simulator does not import the analytic predictor or its state-transition
implementation.

Monte Carlo sampling is increased adaptively until the $95\%$ confidence
interval half-width is at most one quarter of the observed pairwise gap.

All twelve primary frozen cases confirm the optimizer-aware ordering.

The purpose of this protocol is not to make \method an independent predictor in
the exact quadratic regime---Proposition~\ref{prop:exact} already establishes
exactness---but to verify that the identified reversals correspond to genuine
differences under independently implemented dynamics.


\section{Additional Quadratic Results}
\label{app:quadratic}

\subsection{Controlled-grid ranking}

\begin{table}[h]
\centering
\small

\caption{
Controlled-grid mechanism ranking.
\method is exact by construction in this domain.
}
\label{tab:grid_full}

\begin{tabular}{lrrrr}
\toprule
Score
& Pairwise
& Kendall $\tau$
& Rel. regret
& Top-1 \\
\midrule

Static MeanSE
& 58.0\%
& 0.143
& 12.7\%
& 33.3\% \\

LR-aware MF
& 60.8\%
& 0.200
& 11.7\%
& 33.3\% \\

Curv.+sched., $\mu=0$
& 75.8\%
& 0.501
& 20.1\%
& 36.7\% \\

\method
& $100\%^\dagger$
& $1.000^\dagger$
& $0.0\%^\dagger$
& $100\%^\dagger$ \\

\bottomrule
\end{tabular}

\vspace{1mm}

{\footnotesize
$^\dagger$Exact under the modeled linear-quadratic dynamics.
}

\end{table}

\subsection{Mechanism winners across momentum}

Across a broader grid of momentum, schedule, and curvature spectra, the
winning mechanism changes systematically.

Representative exact damage values at moderate ill-conditioning and
$\mu=0.9$ are:

\begin{center}
\begin{tabular}{lrrrr}
\toprule
Schedule
& Dense
& BandMF
& Toeplitz
& Best \\
\midrule
Constant
& 21.86
& 22.12
& 34.26
& Dense \\
Cosine
& 0.58
& 0.51
& 0.44
& Toeplitz \\
Warmup-cosine
& 0.66
& 0.58
& 0.50
& Toeplitz \\
\bottomrule
\end{tabular}
\end{center}

The result is therefore better described as an interaction among temporal
covariance, curvature, momentum, and schedule than as a universal main effect
of momentum.

\subsection{Structured held-outs}

Representative pairwise accuracies for static, LR-aware, and
curvature+schedule-at-zero-momentum scoring are:

\begin{itemize}[leftmargin=1.5em]
    \item held-out momentum:
    $55.0/53.2/60.8\%$;

    \item held-out schedule:
    $58.5/62.8/68.6\%$;

    \item held-out $\kappa=1000$:
    $68.3/70.4/83.9\%$;

    \item held-out BISR:
    $59.1/59.1/73.3\%$.
\end{itemize}

\method is exact in these quadratic held-outs and is therefore not interpreted
as an independent generalization baseline.

\subsection{Curvature perturbation and privacy strength}

Quadratic pairwise rankings remain at $100\%$ through $25\%$ curvature
perturbation in the tested regimes.

At $50\%$ perturbation:

\[
\text{pairwise accuracy}=95.1\%,
\]

\[
\text{top-1 match}=84.4\%,
\]

and

\[
\text{mean relative regret}=1.07\%.
\]

This should not be interpreted as robustness to nonstationary neural
curvature.

Under common sensitivity and Gaussian calibration, changing
$\varepsilon$ rescales exact covariance-induced damage by a common positive
factor, so exact rankings are invariant in this controlled single-participation
setting.


\section{Logistic-Regression Details}
\label{app:logistic}

\subsection{Data generation}

The synthetic logistic-regression task uses

\[
d=16,
\qquad
N=4096.
\]

Let

\[
s
=
\operatorname{logspace}
(-2,-1,16).
\]

Features are sampled independently as

\[
x_i
\sim
\mathcal N
\left(
0,\operatorname{diag}(s)
\right),
\]

and

\[
w^\star
\sim
\mathcal N
\left(
0,I_{16}/16
\right).
\]

Labels are generated according to

\[
p_i
=
\sigma(x_i^\top w^\star),
\qquad
y_i
\sim
\operatorname{Bernoulli}(p_i).
\]

The $4096$ examples are divided into

\[
T=128
\]

disjoint shards of size $32$.

\subsection{Clipping and privacy calibration}

Before private optimization, the clipping threshold is calibrated as the
median per-example gradient norm on the designated calibration split at
initialization.

The threshold is then fixed for the entire private training trajectory.

Headline experiments use

\[
(\varepsilon,\delta)
=
(4,10^{-5})
\]

under single-participation semantics.

\subsection{Curvature estimation}

NoiseCurve uses the normalized feature-only logistic upper bound

\[
H_{\mathrm{cal}}
=
\frac{1}{4N_{\mathrm{cal}}}
X_{\mathrm{cal}}^\top
X_{\mathrm{cal}},
\qquad
N_{\mathrm{cal}}=500.
\]

At $\theta=0$, the logistic weight satisfies

\[
p(1-p)=\frac14,
\]

making this expression exact for the initialization curvature of the empirical
objective on that calibration sample.

\subsection{Curvature-source robustness}

As a post-hoc robustness diagnostic, we also compute

\[
H_{\mathrm{full}}
=
\frac{1}{4N}
X^\top X,
\qquad
N=4096.
\]

The largest relative eigenvalue discrepancy between
$H_{\mathrm{cal}}$ and
$H_{\mathrm{full}}$
is $10.5\%$.

The predicted top-1 mechanism is unchanged in all four optimizer conditions.
The complete ranking is unchanged in three of four.

Kendall correlations with empirical mechanism ordering are

\[
(0.389,0.556,0.444,0.333)
\]

for $H_{\mathrm{cal}}$, and

\[
(0.389,0.556,0.444,0.222)
\]

for $H_{\mathrm{full}}$.

Thus, replacing the calibration curvature by this full-data static estimate
does not explain the nonlinear mismatch.

This check rules out the choice between these two static curvature sources as
the dominant explanation; it does not measure trajectory-wise curvature drift.

The full-training-set estimate is used only as a post-hoc diagnostic and is
not proposed as a privacy-free deployment procedure.

\subsection{Optimizer conditions}

We evaluate:

\begin{itemize}[leftmargin=1.5em]
    \item L0:
    $\mu=0$, constant learning rate;

    \item L1:
    $\mu=0.9$, constant learning rate;

    \item L2:
    $\mu=0.9$, cosine schedule;

    \item L3:
    $\mu=0.9$, warmup-cosine schedule.
\end{itemize}

Banded mechanisms use semi-bandwidth

\[
b=16.
\]

\subsection{Broader ranking evaluation}

Across 61 statistically resolved pairwise comparisons:

\[
\text{Static pairwise accuracy}=91.8\%,
\]

\[
\text{\method pairwise accuracy}=95.1\%.
\]

Mean relative selection regret changes from

\[
4.41\%
\quad\rightarrow\quad
3.53\%.
\]

Top-1 agreement remains

\[
44.4\%
\]

for both methods.

Four of the twelve frozen quadratic crossover cases remain statistically
resolved in logistic regression, and \method predicts all four.


\section{CIFAR-10 CNN Details and Negative Result}
\label{app:cnn}

\subsection{Architecture and data}

The CNN is implemented directly in JAX and contains $140{,}714$ parameters:

\[
\begin{aligned}
&
\operatorname{Conv}
(3\!\rightarrow\!32,3\times3,\mathrm{pad}=1)
+\operatorname{ReLU},
\\
&
\operatorname{Conv}
(32\!\rightarrow\!32,3\times3,\mathrm{pad}=1)
+\operatorname{ReLU},
\\
&
\operatorname{MaxPool}(2\times2),
\\
&
\operatorname{Conv}
(32\!\rightarrow\!64,3\times3,\mathrm{pad}=1)
+\operatorname{ReLU},
\\
&
\operatorname{Conv}
(64\!\rightarrow\!64,3\times3,\mathrm{pad}=1)
+\operatorname{ReLU},
\\
&
\operatorname{MaxPool}(2\times2),
\\
&
\operatorname{Conv}
(64\!\rightarrow\!128,3\times3,\mathrm{pad}=1)
+\operatorname{ReLU},
\\
&
\operatorname{GlobalAvgPool},
\qquad
\operatorname{Dense}(128\!\rightarrow\!10).
\end{aligned}
\]

No batch normalization or weight decay is used.

CIFAR-10 is partitioned into:

\[
40{,}000
\]

private-training examples,

\[
2{,}000
\]

public-calibration examples,

\[
3{,}000
\]

validation examples, and

\[
5{,}000
\]

test examples.

Training uses

\[
T=100
\]

disjoint private shards of size $400$ and five paired seeds.

\subsection{Optimization and clipping}

The per-example clipping threshold is calibrated as the median gradient norm
over 200 public calibration examples and is then fixed during private training.

Curvature is estimated exclusively on the public calibration split using:

\begin{itemize}[leftmargin=1.5em]
    \item Hessian-vector products;
    \item the top-30 Lanczos eigenvalues; and
    \item a Hutchinson trace estimate.
\end{itemize}

An initial configuration with

\[
\eta_0=0.05
\]

produced

\[
\eta_0\lambda_{\max}
\approx6.5
\]

and was rejected before scientific analysis.

The corrected base learning rate is

\[
\eta_0
=
\frac{0.3}{\lambda_{\max}}
\approx0.00229.
\]

The SGD condition uses $\mu=0$.
The three momentum conditions use $\mu=0.9$.

\subsection{Mechanism pool}

The CNN pilot evaluates five mechanism families:

\begin{itemize}[leftmargin=1.5em]
    \item IID Gaussian;
    \item BandMF;
    \item Toeplitz MF;
    \item dense MF; and
    \item learning-rate-aware MF.
\end{itemize}

BandMF and Toeplitz constructions use semi-bandwidth $b=16$.

BISR is not included because the validated strategy available for that pilot
was constructed at horizon $T=128$, while the CNN uses $T=100$.

NoiseCurve is also absent because the CNN experiment was completed before the
independent NoiseCurve reproduction and validation stage.
The pilot was not retroactively rerun solely to enlarge the mechanism pool.

\subsection{Five-seed results}

\begin{table}[h]
\centering
\small

\caption{
CIFAR-10 five-seed pilot after stability calibration.
}
\label{tab:cnn5}

\begin{tabular}{lcccc}
\toprule
Condition
& Best
& Worst
& Spread
& Combined SE \\
\midrule

SGD + constant
& BandMF 12.3\%
& LR-aware 12.0\%
& 0.34 pp
& 0.32 pp \\

Momentum + constant
& IID/Toeplitz 22.7\%
& LR-aware 22.2\%
& 0.48 pp
& 0.34 pp \\

Momentum + cosine
& IID 17.6\%
& Dense 17.3\%
& 0.36 pp
& 0.27 pp \\

Momentum + warmup-cosine
& BandMF 17.1\%
& Toeplitz 16.8\%
& 0.29 pp
& 0.15 pp \\

\bottomrule
\end{tabular}

\end{table}

\subsection{Interpretation of the negative result}

The CNN pilot does not establish a reliable mechanism-ranking signal for the
local \method approximation.

Several properties distinguish this regime from the controlled and logistic
studies:

\begin{itemize}[leftmargin=1.5em]

    \item single-participation semantics restrict the trajectory to only
    $T=100$ updates;

    \item stochastic-gradient variability introduces an additional source of
    optimization noise;

    \item per-example clipping changes gradient magnitude and direction;

    \item curvature changes along the trajectory; and

    \item different privacy-noise realizations induce different optimization
    trajectories.

\end{itemize}

The parameter dimension is also much larger:

\[
d=140{,}714
\]

versus $d=16$ in the controlled and logistic studies.

For isotropic Gaussian perturbations, expected squared noise energy scales
with dimension, which may make covariance-specific differences harder to
resolve.
We do not isolate this effect experimentally and therefore treat it only as a
possible contributor.

The observed best--worst differences are small and directionally inconsistent,
so the experiment was not expanded to additional seeds solely to search for
statistical significance.

Accordingly, the CNN result is an empirical boundary of the current
fixed-curvature local approximation rather than evidence that optimizer state
is irrelevant to neural private training.


\section{Extended Discussion and Limitations}
\label{app:impact}

\subsection{Evaluation versus covariance synthesis}

The present work primarily studies \emph{mechanism evaluation and selection}.

Given a collection of privacy-valid temporal covariances, \method quantifies
how each covariance is filtered by the specified optimizer dynamics.

The transfer representation also suggests a constructive extension:
one could directly optimize covariance structure against optimizer-aware
privacy damage while imposing the sensitivity, positive-definiteness,
participation, and structural constraints required by the matrix mechanism.

We deliberately separate evaluation from synthesis.
Establishing that optimizer state changes mechanism selection does not require
claiming that the current paper constructs the globally optimal
momentum-specific covariance.

\subsection{Why neural transfer is harder}

The strongest agreement with \method occurs where the transfer model is exact:
fixed quadratic dynamics.

Private logistic regression shows that the optimizer-state distinction can
persist outside that domain, but the transfer is attenuated.

In neural training, stochastic-gradient variability, clipping, curvature
drift, and mechanism-dependent trajectory feedback become simultaneously
important.

A useful future direction is to explicitly decompose optimization variability
into privacy-noise, minibatch-sampling, and clipping-induced components and
identify regimes in which optimizer-filtered privacy noise is sufficiently
large to influence mechanism selection.

\subsection{Trajectory feedback and nonstationary curvature}

The transfer operator in the present paper is local.

In nonlinear optimization,

\[
z_t
\longrightarrow
\theta_{t+1}
\longrightarrow
\nabla f(\theta_{t+1}),
H(\theta_{t+1})
\longrightarrow
\text{future optimizer response}.
\]

Different privacy covariances can therefore induce different optimization
trajectories and different sequences of effective curvature.

A single fixed transfer operator cannot represent this feedback exactly.

The logistic curvature-source diagnostic shows that replacing one static
curvature estimate by a larger-sample static estimate does not resolve the
observed disagreement.
It does not directly measure temporal curvature drift.

\subsection{Adaptive optimizers}

Persistent optimizer state under differential privacy has also been studied
from other perspectives.

DiSK uses Kalman filtering to denoise privatized gradients
\citep{zhang2025disk}, while FIBER studies interactions between temporal
filtering and AdamW's adaptive second-moment state
\citep{dm2026fiber}.

These approaches modify the private optimizer or its internal filtering
procedure.

\method addresses the complementary question of how a fixed privacy-valid
temporal covariance is filtered by the optimizer and how that interaction
changes mechanism ranking.

Extending the present exact formulation to Adam or AdamW would require
representing additional first- and second-moment states and nonlinear
coordinate-wise normalization.

\subsection{Repeated participation and privacy amplification}

The cleanest headline comparisons use single-participation semantics.

In repeated-participation, subsampling, shuffling, or amplification regimes,
candidate mechanisms may differ not only in covariance structure but also in
effective sensitivity and privacy calibration.

The observation that a common scalar Gaussian scale leaves exact quadratic
rankings unchanged therefore does not automatically extend to these regimes.

Optimizer-aware utility should be considered only after privacy validity and
calibration have been established under the actual participation model.

\subsection{Practical implications}

A practical implication of the results is that privacy mechanism and optimizer
configuration need not be independent design choices.

A covariance that performs well under one optimizer regime can become less
favorable after introducing momentum even when its differential-privacy
guarantee remains unchanged.

The work does not establish accelerator-level speedups, lower hardware cost,
or production-scale performance gains.

\method operates at the level of optimization dynamics and temporal covariance
and is therefore hardware agnostic.

\subsection{Curvature access and privacy}

Curvature information can itself constitute a privacy boundary.

If Hessians, Hessian-vector products, or related statistics are computed from
protected examples, that access may require its own privacy treatment.

The CNN experiment uses public calibration data for curvature estimation.

For logistic regression, the full-training-data curvature evaluation described
in Appendix~\ref{app:logistic} is used only as a post-hoc diagnostic rather
than as part of the privacy-compatible training procedure.


\section{Ethics, AI Use, and Reproducibility}
\label{app:repro}

\subsection{Ethics}

This work studies mechanisms for differentially private training and does not
involve human-subject experiments.

Its intended impact is to improve the reliability and utility of
privacy-preserving optimization.

Utility comparisons are meaningful only after privacy semantics and
sensitivity are matched.
Any protected-data auxiliary computation, including curvature estimation,
would need to be incorporated into the relevant privacy analysis.

\subsection{AI use}

Generative AI tools were used for language editing and organization of
manuscript drafts, assistance in formulating experimental checklists, and
assistance with code drafting/debugging.

All mathematical claims, literature relationships, experimental
implementations, numerical results, and interpretations reported in the paper
were reviewed and validated by the authors.

The authors take responsibility for the final content.

\subsection{Reproducibility}

The intended anonymized artifact includes:

\begin{enumerate}[leftmargin=1.5em]

    \item environment and version manifests;

    \item mechanism sensitivity checks;

    \item learning-rate-aware and NoiseCurve reproduction tests;

    \item the 576-case identity grid;

    \item the frozen crossover registry and SHA-256 hash;

    \item independent NumPy validation code;

    \item logistic and CNN experiment manifests; and

    \item scripts reproducing reported tables.

\end{enumerate}

Random sources for data generation, initialization, privacy noise, mechanism
optimization, and replication are separated.
\end{appendices}

\end{document}